\documentclass[10pt,twocolumn,letterpaper]{article}
\usepackage{times}
\usepackage{graphicx,amsmath,amssymb,booktabs}
\usepackage[dvipsnames]{xcolor}
\usepackage[numbers,sort&compress]{natbib}
\usepackage[hyphens]{url}
\usepackage[format=plain,labelformat=simple,labelsep=period,font=small,compatibility=false]{caption}
\usepackage[shortlabels,inline]{enumitem}
\setlist[itemize]{noitemsep,leftmargin=*,topsep=0em}
\setlist[enumerate]{noitemsep,leftmargin=*,topsep=0em}
\makeatletter
\renewcommand{\section}{\@startsection{section}{1}{\z@}{-10pt plus -2pt minus -2pt}{7pt}{\large\bfseries}}
\renewcommand{\subsection}{\@startsection{subsection}{2}{\z@}{-8pt plus -2pt minus -2pt}{5pt}{\fontsize{11}{13}\selectfont\bfseries}}
\renewcommand{\subsubsection}{\@startsection{subsubsection}{3}{\z@}{-6pt plus -2pt minus -2pt}{3pt}{\normalsize\bfseries}}
\renewcommand{\@seccntformat}[1]{\csname the#1\endcsname.\ }
\makeatother

\usepackage[T1]{fontenc}
\usepackage[utf8]{inputenc}
\usepackage{placeins}
\usepackage{needspace}
\usepackage{stfloats}
\usepackage{float}
\definecolor{linkblue}{rgb}{0.21,0.49,0.74}
\usepackage[breaklinks,colorlinks,allcolors=linkblue]{hyperref}
\hypersetup{
  pdftitle={ReRead: Wrong Predictions, Useful Representations - Correcting Frozen Robot Policies by Re-reading Their Own Representations},
  pdfauthor={Xu Su, Shunlong Xiao, Qintao Yi, Mingkuan Zhao, Xinmiao Yu, Yizhe Li, Zihan Wang, Bokui Chen}
}
\title{ReRead: Wrong Predictions, Useful Representations\newline Correcting Frozen Robot Policies by Re-reading Their Own Representations}
\author{
Xu Su$^{1}$\quad Shunlong Xiao$^{1}$\quad Qintao Yi$^{1}$\quad Mingkuan Zhao$^{2}$\\
Xinmiao Yu$^{1}$\quad Yizhe Li$^{1}$\quad Zihan Wang$^{1}$\quad Bokui Chen$^{1}$\\[4pt]
{\normalsize $^{1}$Tsinghua University\qquad $^{2}$Xi'an Jiaotong University}
}
\date{}
\begin{document}
\onecolumn
\makeatletter
\renewcommand{\@maketitle}{%
  \noindent\begin{minipage}{\textwidth}
    \centering
    {\Large\bfseries\@title\par}
    \vspace{6pt}
    {\large\begin{tabular}[t]{c}\@author\end{tabular}\par}
    \vspace{6pt}
    \begin{minipage}{\textwidth}
  \centering
  \captionsetup{type=figure}
  \includegraphics[width=0.91\textwidth]{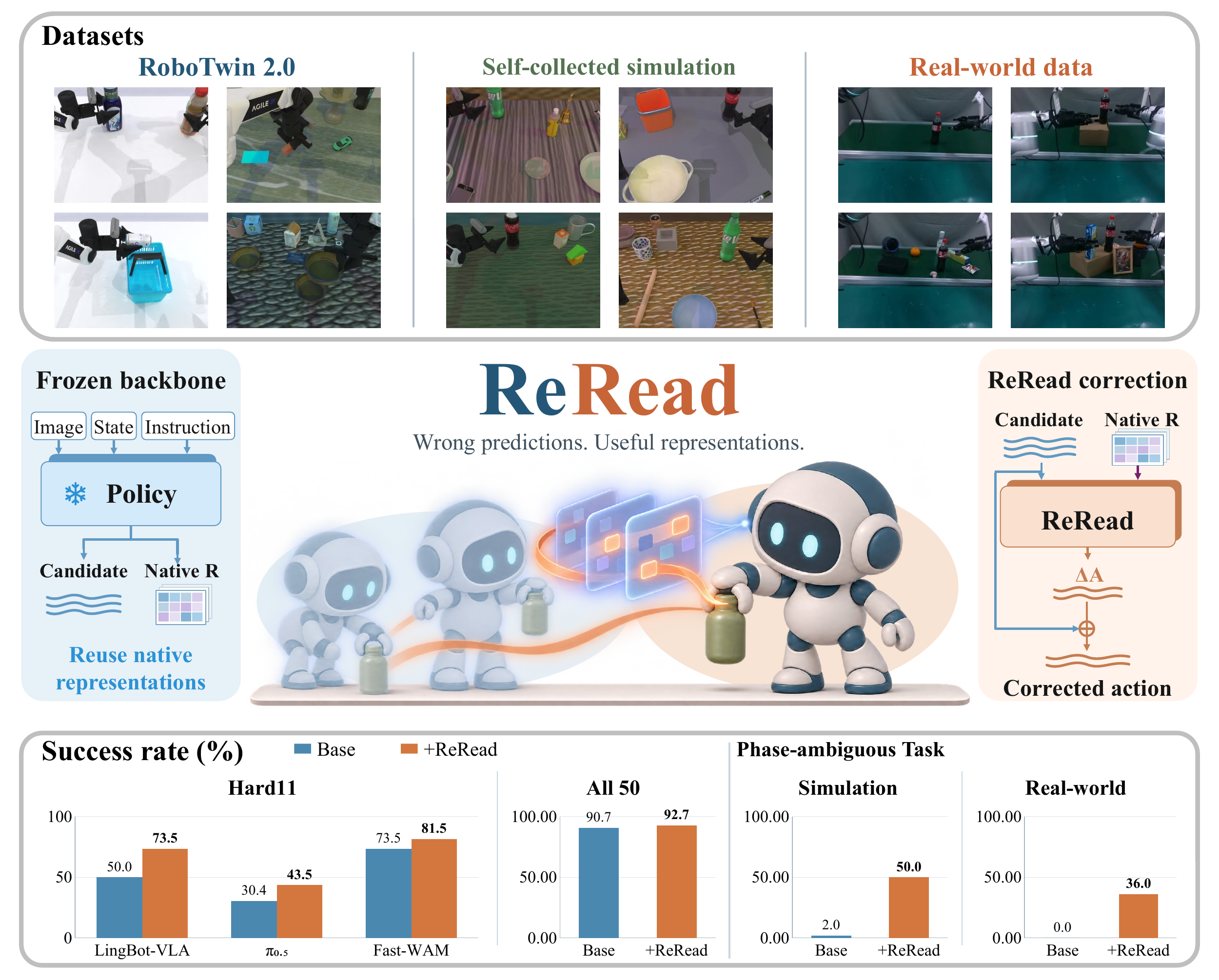}
  \caption{Overview of ReRead.}
  \label{fig:reread_teaser}
\end{minipage}
\par\vspace{6pt}

    \begin{minipage}{\textwidth}
      \renewcommand{\abstract}{\centerline{\large\bfseries Abstract}\vspace{6pt}\noindent\itshape\ignorespaces}
      \renewcommand{\endabstract}{\par}
      \begin{abstract}
When a pretrained robot policy fails on challenging tasks, common remedies include retraining the backbone, adding an external visual encoder, or introducing a memory module. We argue that wrong predictions do not imply wrong representations: a policy can retain the required information yet fail to read it out as the correct action. To test this view, we propose ReRead: with the backbone frozen, a corrector re-reads the current and two previous caches to predict an action residual. The corrector sees no images, is trained solely on offline demonstrations from all 50 RoboTwin 2.0 tasks, and adds only 4.17\% latency. Ablations and interventions show that the correction relies on the backbone's own representations: none of four external visual encoders adds further gain, and swapping in past caches from other trajectories sharply increases action error. On the 11 hardest RoboTwin 2.0 tasks, ReRead improves the success rates of LingBot-VLA, $\pi_{0.5}$, and Fast-WAM by 23.45, 13.09, and 8.00 percentage points, respectively. In paired evaluation on all 50 tasks, it raises Fast-WAM's average success rate from 90.73\% to 92.66\%. On a Flexiv dual-arm robot, it raises success on a phase-ambiguous task from 0\% to 36\%.
\end{abstract}

    \end{minipage}
  \end{minipage}
}
\makeatother
\maketitle
\twocolumn
\section{Introduction}
\label{sec:intro}
When a pretrained robot policy fails on challenging tasks, common remedies include retraining the backbone, adding an external visual encoder, or introducing a memory module~\cite{octo2024,memoryvla2026,hamlet2026}.
These remedies share an implicit assumption: if the predicted actions are wrong, the policy's representations must lack task-relevant information.

But failure has another possible cause: the policy already has the information it needs, yet fails to read it out as the correct action.
Similar findings in image recognition show that retraining only the classifier can improve accuracy while keeping the learned representations fixed~\cite{kang2020decoupling,kirichenko2023dfr}.
Robot tasks have a further property: the current frame is often insufficient to determine the action, because the action also depends on the current task phase.
The information needed to distinguish phases is not in the current frame, but it may already reside in the internal representations of the policy's previous decisions, which are discarded once the actions are output.

We therefore argue that \textbf{wrong predictions do not imply wrong representations}.
The attention keys and values (K/V) cached by the backbone, the robot state, and temporal information form an internal memory, and we call its task-relevant information the \emph{representation}.
(In Figs.~\ref{fig:reread_teaser} and~\ref{fig:newm_overview}, ``R'' stands for ``Representations''.)
The process that maps the internal memory to an action is the \emph{readout}, and the action finally returned is the \emph{prediction}.
A backbone error may arise because the required information was never in the caches, or because it is in the caches but is not read out as the correct action.
In the latter case, it is the readout that needs to be improved, and the backbone need not be retrained.
This corresponds to the classical decomposition under squared loss: the prediction error equals the representation error plus the readout gap.

To test this view, we propose ReRead (Fig.~\ref{fig:reread_teaser}): with the backbone frozen, a new corrector re-reads the representations cached by the backbone.
The backbone generates candidate actions as usual, and ReRead stores the native visual and action K/V from each of its decisions.
Using the candidate action as the query, the corrector reads the caches of the current and the two previous decisions and outputs an action residual, which is added back to the candidate action.
The corrector sees no images, does not retrain the backbone, and does not re-encode the history.
It is trained solely on offline demonstrations with an action MSE loss and adds 11.66 ms of latency per decision (4.17\%).

We train ReRead on offline demonstrations from all 50 tasks of RoboTwin 2.0~\cite{robotwin2_2026}.
Fast-WAM~\cite{fastwam2026} has a reported average success rate of 91.83\% on these tasks, but its failures are concentrated in a few tasks: the 11 tasks with the lowest published success rates (Hard11) account for about 66.5\% of the failures.
We use Hard11 as an evaluation subset to analyze cases in which the policy retains the required information but predicts the wrong action, and evaluate the overall effect of re-reading on all 50 tasks.

The experiments are organized around four questions.

First, when the backbone errs, is the required information still in the caches? Re-reading the current cache alone reduces the native candidate's open-loop MSE by 61.45\%. With ReRead, the success rates of LingBot-VLA, $\pi_{0.5}$, and Fast-WAM on Hard11 improve by 23.45, 13.09, and 8.00 percentage points, respectively. In our paired evaluation on all 50 tasks, Fast-WAM's average success rate rises from 90.73\% to 92.66\%.

Second, can prediction and representation be evaluated separately? Intervention experiments show that the corrected action is determined mainly by the caches: increasing the error of the candidate action 2-fold increases the error of the corrected action by only 12.08\%, whereas replacing the past caches with those from other trajectories of the same task increases it 12.37-fold.

Third, do backbone failures stem from insufficient perception? None of the four external visual encoders outperforms ReRead, whereas replacing the past caches markedly changes the corrected action. These results do not support the perception-bottleneck explanation.

Fourth, when the current frame cannot distinguish phases, where is the required information? On a phase-ambiguous task, ReRead raises the success rate from 2\% to 50\% in simulation and from 0\% to 36\% on a real Flexiv dual-arm robot. Swapping in caches from the opposite phase moves the corrected action toward the other phase, indicating that the required information lies in the caches of previous decisions.

Our contributions are as follows:
\begin{itemize}
    \item[\textbf{\textbullet}] We propose a new perspective: wrong predictions do not imply wrong representations. We decompose the prediction error into the representation error and the readout gap, and estimate the part that re-reading can eliminate.
    \item[\textbf{\textbullet}] To test this perspective, we propose ReRead, which freezes the backbone and trains a corrector solely on offline demonstrations to re-read the backbone's cached representations and correct the actions. Our training and inference code and model weights are publicly available at \url{https://github.com/tan2355/reread}.
    \item[\textbf{\textbullet}] Interventions and ablations show that the correction relies on the backbone's own caches. Phase information also remains in the caches of previous decisions. Prediction and representation can therefore be evaluated separately.
\end{itemize}

\section{Related Work}
\label{sec:background}
\paragraph{Gap between Representations and Outputs.}
Good internal representations do not imply good outputs.
In image recognition, the representation and the classifier can be adjusted separately for long-tailed recognition~\cite{kang2020decoupling}, and the representations of models that rely on spurious correlations still contain robust features~\cite{kirichenko2023dfr}.
In language models, truth signals that do not depend on the surface form of the answer can be read out from internal activations~\cite{burns2023ccs}, and simply intervening on attention-head activations improves answer truthfulness~\cite{li2023iti}.
Probing studies examine both how much information can be read out from representations~\cite{pimentel2020probing} and how much these readouts depend on the capacity of the probe itself~\cite{hewitt2019control}.
Unlike these works, we focus on the continuous actions of frozen robot policies, where the information is also spread across the past few decisions, and we directly quantify how much error re-reading can eliminate.
\paragraph{Temporal Memory in Robot Policies.}
History helps policies distinguish visually similar task phases.
Octo processes recent observations~\cite{octo2024}.
MemoryVLA and HAMLET represent, retrieve, or compress history to condition action generation~\cite{memoryvla2026,hamlet2026}.
GeoMem-VLA stores geometric and observation representations in temporal memory and retrieves them to guide action generation~\cite{geomemvla2026}.
HyVLA-0.5 compresses visual history through temporal-spatial attention before passing the resulting representations to the VLM~\cite{hyvla2026}.
These methods use history to condition action generation.
If the error lies in the readout, adding historical context does not necessarily help.
In contrast, ReRead leaves the backbone's inputs and candidate generation unchanged and, once the candidate action is produced, uses a corrector to read the current and past caches.
\paragraph{Residual Correction for Frozen Policies.}
Residual policies add a small adjustment on top of existing actions while preserving the backbone's original capabilities.
Residual Policy Learning and FISH use reinforcement learning to optimize a residual on top of the backbone~\cite{silver2018rpl,fish2023}.
Policy Decorator freezes a large imitation policy and learns a small residual through online interaction~\cite{yuan2025policydecorator}.
Dejavu freezes a VLA, retrieves information from an experience database, and learns action corrections from interaction rewards~\cite{dejavu2026}.
Within a framework of multi-round human corrections, I3L also evaluates a supervised residual correction scheme~\cite{i3l2026}.
These methods all rely on online interaction, rewards, or additional human-correction data.
ReRead uses only existing offline demonstrations and reads the backbone's caches from the current episode to test whether they retain the information needed for correct actions.

\section{Separating Prediction from Representation}
\label{sec:theory}
\begin{figure*}[!tbp]
    \centering
    \includegraphics[width=\textwidth]{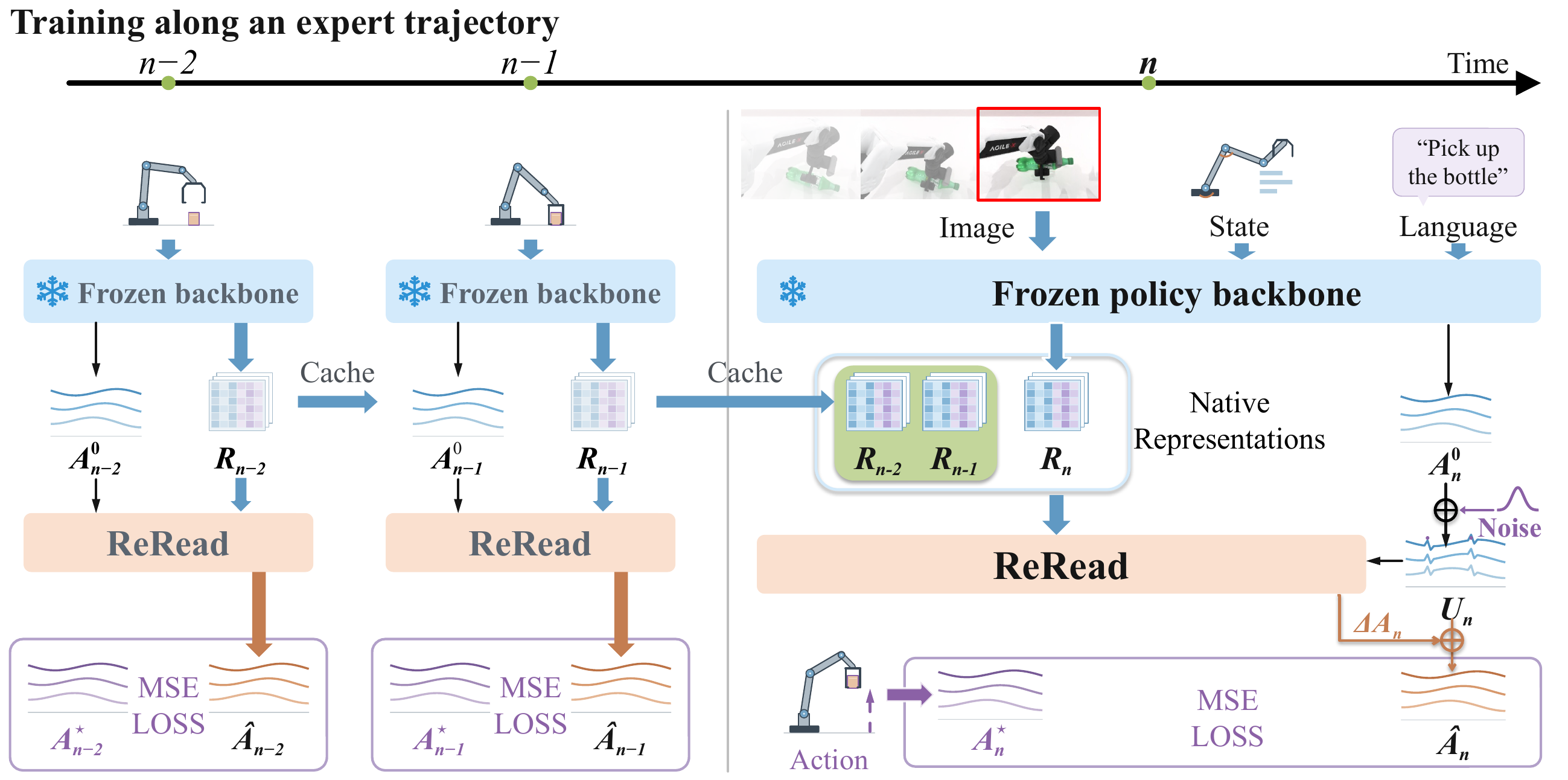}
    \caption{Illustration of ReRead training along expert trajectories.}
    \label{fig:newm_overview}
\end{figure*}
Whether an action prediction is wrong and whether the caches contain useful information are two different questions.
We distinguish them by decomposing the prediction error into the representation error and the readout gap.
\subsection{Problem Formulation}
\label{sec:preliminaries}
At the $n$-th decision, the frozen backbone receives images $I_n$, the robot state $s_n$, and the instruction $\ell$, denoted $x_n=(I_n,s_n,\ell)$, and generates a candidate action chunk from sampling noise $\xi_n$
\begin{equation}
    A_n^0=\pi_\theta(x_n;\xi_n)\in\mathbb R^{H\times d},
    \label{eq:base_candidate}
\end{equation}
where $H$ is the chunk length and $d$ is the action dimension.
The native K/V, state, and temporal information produced by this decision together form a cache. ReRead combines the current and past caches into a memory $\mathcal M_n$ and learns a residual correction
\begin{equation}
    \widehat A_n=A_n^0+R_\phi(A_n^0,\mathcal M_n),
    \label{eq:residual_inference}
\end{equation}
supervised by the expert action $A_n^\star$, while the backbone parameters $\theta$ remain frozen.
\subsection{Two Sources of Prediction Error}
\label{sec:correction_theory}
In what follows, we fix the backbone and the evaluation distribution and omit the subscript $n$.
We denote the corrector's input by $Z=(A^0,\mathcal M)$ and its output residual by $r_\phi(Z)=R_\phi(A^0,\mathcal M)$.
The mean squared error after adding a residual $r$ is $\mathcal R(r)=\mathbb E\|A^\star-A^0-r(Z)\|^2$, where $\mathcal R(0)$ is the error of executing the candidate directly.
Under squared loss, the best prediction from $Z$ is the conditional mean $m_Z=\mathbb E[A^\star\mid Z]$.
By the projection property of the conditional mean~\cite{hastie2009esl}, we have
\begin{equation}
\begin{aligned}
    \mathcal R(0)
       &=\underbrace{\mathbb E\|A^\star-m_Z\|^2}_{\mathcal K(Z):\,\text{representation error}}\\
       &\quad+\underbrace{\mathbb E\|m_Z-A^0\|^2}_{G(Z):\,\text{readout gap}}.
\end{aligned}
\label{eq:correctable_decomposition}
\end{equation}
The first term is the representation error: the minimum prediction error given the candidate action and the caches.
Hence, even when the prediction error is large, the representation error can still be small; the optimal residual $r^\star(Z)=m_Z-A^0$ exactly eliminates the entire readout gap.

The optimal readout is not directly available, but its prediction error can be bounded above with a trained corrector. When $\mathcal R(0)>0$,
\begin{equation}
\begin{aligned}
    b_\phi(Z)&=1-\frac{\mathcal R(r_\phi)}{\mathcal R(0)}
                   \leq\frac{G(Z)}{\mathcal R(0)},\\
    \frac{\mathcal K(Z)}{\mathcal R(0)}&\leq 1-b_\phi(Z).
\end{aligned}
\label{eq:readout_certificate}
\end{equation}
Here, $b_\phi(Z)$ is the fraction of the error actually reduced by the corrector.
The eliminated error cannot exceed the readout gap, and the ratio of the representation error to the initial error is at most $1-b_\phi$.
In the experiments, we use held-out data to estimate both $b_\phi$ and the corresponding upper bound on the representation error.
See Appendix~\ref*{sec:appendix_theory} for the proofs and measurement definitions.
\subsection{Roles of the Candidate and Past Caches}
\label{sec:theory_information_paths}
\paragraph{Candidate changes and cache reading.}
Suppose the candidate is replaced with an arbitrary $U$, while the memory $\mathcal M$ stays unchanged.
If, given the memory, the candidate provides no additional information about the target action, i.e., $\mathbb E[A^\star\mid U,\mathcal M]=\mathbb E[A^\star\mid\mathcal M]=h(\mathcal M)$, then
\begin{equation}
    r^\star(U,\mathcal M)=h(\mathcal M)-U,\qquad
    f^\star(U,\mathcal M)=h(\mathcal M),
    \label{eq:record_sufficient_output}
\end{equation}
where $f^\star=U+r^\star$ is the optimal corrected action.
That is, the optimal residual exactly cancels the change in the candidate, and the final action is determined solely by the caches.
\paragraph{Why history remains useful.}
Let $Z_c$ denote the current candidate and the current cache, let $Z_m=(Z_c,\mathcal H^-)$ denote the input that further includes the ordered history, and let $m_i=\mathbb E[A^\star\mid Z_i]$.
After adding the history, the optimal error decreases by
\begin{equation}
    G_{\rm hist}=\mathcal K(Z_c)-\mathcal K(Z_m)
       =\mathbb E\|m_m-m_c\|^2.
\label{eq:history_information_gain}
\end{equation}
Thus, the gain is positive as long as the history changes the conditional mean of the target action.
For example, two phases with similar current frames but different target actions can be distinguished using the past caches.
This gain does not require the past action predictions to be correct.
Appendix~\ref*{sec:appendix_theory_example} constructs an example in which all past predictions are wrong, yet the caches still recover the phase.

\section{ReRead: Re-reading the Policy's Cache}
\label{sec:method}
\subsection{Native Caches}
\label{sec:memory}
After each decision, ReRead (Fig.~\ref{fig:newm_overview}) directly stores the native K/V computed by the backbone, rather than re-encoding the observations with an external encoder. Each cache record contains K/V from multiple layers, the robot state, the decision anchor (the time step on the demonstration trajectory at which a new decision is made, once every $S=20$ executed steps), and the executed prefix. The visual K/V come from the prefill of the current observation, and the action K/V, which come from the last denoising call, are the backbone's internal plan when generating the candidate action. By default, the corrector reads the caches of the current and the two previous decisions (corresponding to $t_n,t_n-S,t_n-2S$) and receives no images; all the scene content it sees is computed by the backbone.

For cache $\tau$, source layer $l$, and token position $i$, we construct
\begin{equation}
    B_{\tau li}=\operatorname{LN}\!\left(
       W_K\operatorname{sg}(K_{\tau li})
      +W_V\operatorname{sg}(V_{\tau li})+W_s s_\tau
    \right),
    \label{eq:memory_content}
\end{equation}
where $\operatorname{LN}$ denotes layer normalization, $\operatorname{sg}$ denotes stop-gradient, and $W_K,W_V,W_s$ project the frozen K/V and the state into the corrector's space. The positional encoding $P_{\tau li}$ marks the source layer, camera, modality, and spatial or action position of each token, and also includes the cache age and the executed prefix. When reading, $B+P$ serves as the keys and $B$ as the values, so matching considers both content and position, while only content is retrieved.
\subsection{Candidate Queries and Residual Correction}
\label{sec:reader}
The candidate action is the plan that the backbone intends to execute, and the corrector uses it as the query to decide what to read from the caches. Specifically, it directly uses the full candidate action values and embeds each action position as a query
\begin{equation}
    q_0=W_a U_n+E^{\rm act}.
\end{equation}
Here, $E^{\rm act}$ denotes the action-position embedding. At inference time, $U_n=A_n^0$; during training, $U_n$ is a perturbed candidate (see Sec.~\ref{sec:supervision}). The corrector has five layers, each applying query self-attention, cross-attention to the caches, and a feed-forward network in turn, all with pre-norm residual connections:
\begin{align}
    q'_j &= q_j+\operatorname{SA}_j(\operatorname{LN}(q_j)),\\
    q''_j &= q'_j+\operatorname{CA}_j
      (\operatorname{LN}(q'_j),B+P,B;M),\\
    q_{j+1} &= q''_j+\operatorname{FFN}_j(\operatorname{LN}(q''_j)),
    \label{eq:reader}
\end{align}
where the mask $M$ masks out missing past caches and those dropped during training; the layers do not share attention parameters. The final output is
\begin{equation}
    \widehat A_n=U_n+W_o\operatorname{LN}(q_5).
    \label{eq:endpoint}
\end{equation}
The output head is zero-initialized, so the initial residual is zero and the corrected action equals the input candidate. The corrector has 198.52M parameters and adds 11.66 ms per decision; see Appendix~\ref*{sec:appendix_resource_cost} for cost measurements.
\subsection{Supervised Training}
\label{sec:supervision}
\paragraph{Candidate perturbation.}
If the candidate always remained unchanged during training, the corrector could reduce the error merely by calibrating the candidate, without reading the caches. Therefore, during training, we add sparse noise only to the candidate action values provided to the corrector:
\begin{equation}
\begin{split}
    U_{n,h,j}&=A^0_{n,h,j}+v_{n,h}m_{n,h}\epsilon_{n,h,j},\\
    m_{n,h}&\sim\operatorname{Bernoulli}(0.2),\quad
    \epsilon_{n,h,j}\sim\mathcal N(0,0.05^2),
\end{split}
\label{eq:corruption}
\end{equation}
where each action position is selected with probability 0.2, and Gaussian noise with a standard deviation of 0.05 is added to the arm joints at the selected positions (see Appendix~\ref*{sec:appendix_impl}); $v_{n,h}$ indicates whether the label at that position is valid. The backbone inputs, the native K/V, and the expert targets all remain unchanged. The corrector receives only the perturbed candidate and the backbone's caches, without direct access to environment observations. By Eq.~\eqref{eq:record_sufficient_output}, when the caches suffice to determine the target action, the optimal residual exactly cancels the perturbation.
\paragraph{Training objective.}
The supervision target $A_n^\star$ is the original continuous action chunk at the decision anchor, and training uses only the action MSE:
\begin{equation}
    \mathcal L_{\rm act}=
    \frac{\sum_{b,h,j}v_{b,h}
       (\widehat A_{b,h,j}-A^\star_{b,h,j})^2}
         {d\sum_{b,h}v_{b,h}}.
    \label{eq:objective}
\end{equation}

\paragraph{Causal caching.}
\label{sec:causal}
Training and deployment share the same causal path: the corrector can only read the current and past caches.
During training, we step through each demonstration trajectory in the order of its decision anchors, run the backbone only once at each anchor, and store the resulting native K/V in the cache, where later anchors directly reuse them as history (Fig.~\ref{fig:newm_overview}).
Thus, each training sample requires only one backbone forward pass, the backbone's computation does not grow with the number of past caches, and gradients do not backpropagate across decisions.
The K/V in the cache remain unchanged, whereas the content projection and age encoding are recomputed at every read with the latest corrector parameters.

\section{Experiments}
\label{sec:experiments}
\subsection{Experimental Setup}
\label{sec:exp_setup}
\paragraph{Tasks and metrics.}
We evaluate on the 50 tasks of RoboTwin 2.0~\cite{robotwin2_2026}, with 100 episodes per task in each of the clean and random settings. On Hard11, we evaluate 25 episodes per task and setting, 550 episodes in total. We re-evaluate both the backbones and ReRead on Hard11 under this protocol and report the success rates.
\paragraph{Training.}
On the RoboTwin 2.0 benchmark, we train all correctors from scratch using offline demonstrations from all 50 tasks, with the backbones frozen. We use the same demonstrations and splits for all correctors and train for 20K steps with a global batch size of 1024, using only the action MSE as the objective. Both training and inference use $D=20$ denoising steps and replan after executing $S$ steps, with chunk length $H=32$ and action dimension $d=14$. Each ablation variant is trained separately. For Fast-WAM, the Hard11 and full-benchmark evaluations use the same ReRead weights. See Appendix~\ref*{sec:appendix_impl} for implementation details and resource measurements.
\paragraph{Open-loop measurement.}
The open-loop analysis fixes the trained ReRead and computes statistics over the first $S$ executed steps on the 22 held-out expert trajectories of Hard11. The past K/V come from the backbone's forward passes when it makes its own predictions.

\begin{figure}[!htbp]
    \centering
    \includegraphics[width=\linewidth]{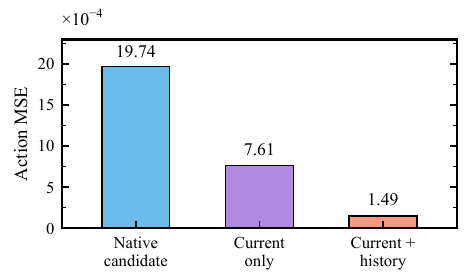}
    \caption{Open-loop readout error.}
    \label{fig:readout_gap}
\end{figure}

\subsection{Is the Required Information Still in the Caches?}
\label{sec:exp_readout_gap}
\paragraph{Open loop: readout gap.}
Re-reading the current cache alone reduces the native candidate's MSE from $19.74\times10^{-4}$ to $7.61\times10^{-4}$, a 61.45\% reduction. Adding the caches of the two previous decisions further reduces it to $1.49\times10^{-4}$, a total reduction of 92.45\% (Fig.~\ref{fig:readout_gap}). The MSE decreases relative to the native candidate on all 11 tasks, and on 10 of them it decreases further once the past caches are added. We thus estimate that the representation error accounts for at most about 7.55\% of the initial error. The candidate sampling variance accounts for only 4.20\% of the prediction error, and directly averaging six candidates reduces the MSE by only 3.52\%, far less than the reduction from reading the caches (Appendix~\ref*{sec:appendix_open_loop}).

\paragraph{Closed loop: three backbones.}
We attach the corrector to LingBot-VLA, $\pi_{0.5}$, and Fast-WAM, which differ in architecture and include two VLAs and one video world action model. ReRead improves their success rates on Hard11 by 23.45, 13.09, and 8.00 percentage points, respectively (Tab.~\ref{tab:local_backbones}). See Appendix~\ref{sec:appendix_local_backbones} (Tab.~\ref{tab:local_backbone_tasks}) for per-task results.

\begin{table}[!htbp]
    \centering
    \caption{Closed-loop results on Hard11 across three backbones (local re-evaluation).}
    \label{tab:local_backbones}
    \small
    \setlength{\tabcolsep}{2.0pt}
    \begin{tabular}{@{}lrrr@{}}
        \toprule
        Model & Clean$\uparrow$ & Random$\uparrow$ & Average$\uparrow$\\
        \midrule
        LingBot-VLA~\cite{lingbotvla2026} & 46.55 & 53.45 & 50.00\\
        ReRead-LingBot-VLA & \textbf{75.27} & \textbf{71.64} & \textbf{73.45}\\
        \midrule
        $\pi_{0.5}$~\cite{motus2025} & 30.18 & 30.55 & 30.36\\
        ReRead-$\pi_{0.5}$ & \textbf{45.45} & \textbf{41.45} & \textbf{43.45}\\
        \midrule
        Fast-WAM~\cite{fastwam2026} & 76.73 & 70.18 & 73.45\\
        ReRead & \textbf{85.45} & \textbf{77.45} & \textbf{81.45}\\
        \bottomrule
    \end{tabular}
\end{table}

\paragraph{Full benchmark.}
Table~\ref{tab:overall} reports the effect of re-reading alongside results from other models trained on RoboTwin 2.0 demonstrations. In our paired evaluation on all 50 tasks, Fast-WAM achieves an average success rate of 90.73\% (Fast-WAM$^{\dagger}$), and ReRead raises it to 92.66\%. See Appendix~\ref*{sec:appendix_closed_loop} for per-task results.

\begin{table}[!htbp]
    \centering
    \caption{Results on all 50 tasks of RoboTwin 2.0. $^{\dagger}$: local re-evaluation; other baselines are published values.}
    \label{tab:overall}
    \small
    \setlength{\tabcolsep}{5pt}
    \begin{tabular}{@{}lrrr@{}}
        \toprule
        Model & Clean$\uparrow$ & Random$\uparrow$ & Average$\uparrow$\\
        \midrule
        $\pi_{0.5}$~\cite{motus2025} & 42.98 & 43.84 & 43.41\\
        LingBot-VLA~\cite{lingbotvla2026} & 86.50 & 85.34 & 85.92\\
        Motus~\cite{motus2025} & 88.66 & 87.02 & 87.84\\
        HyVLA-0.5~\cite{hyvla2026} & 90.90 & 90.06 & 90.48\\
        GeoMem-VLA~\cite{geomemvla2026} & 91.10 & 91.24 & 91.17\\
        LingBot-VA~\cite{lingbotva2026} & 92.90 & 91.50 & 92.20\\
        Fast-WAM~\cite{fastwam2026} & 91.88 & 91.78 & 91.83\\
        Fast-WAM$^{\dagger}$~\cite{fastwam2026} & 91.06 & 90.40 & 90.73\\
        \midrule
        ReRead (Ours) & \textbf{93.20} & \textbf{92.12} & \textbf{92.66}\\
        \bottomrule
    \end{tabular}
\end{table}

\subsection{Are Prediction and Representation Separable?}
\label{sec:exp_separation}
\paragraph{Bidirectional interventions.}
If the corrected action is determined by the caches, it should remain nearly unchanged when only the candidate changes, but change markedly when only the caches change. We fix the trained ReRead and apply two opposite interventions on the same trajectories: one keeps the caches unchanged and adds sparse Gaussian noise to the candidate; the other keeps the candidate and the current K/V unchanged and replaces only the caches of the two previous decisions (design in Appendix~\ref*{sec:appendix_open_loop}; results in Figs.~\ref{fig:candidate_perturbation} and~\ref{fig:history_replacement}).
When only the candidate is changed, noise with $\sigma=0.10$ increases the candidate MSE 2-fold, whereas the MSE of the corrected action increases by only 12.08\%.

\begin{figure}[!htbp]
    \centering
    \includegraphics[scale=1.15]{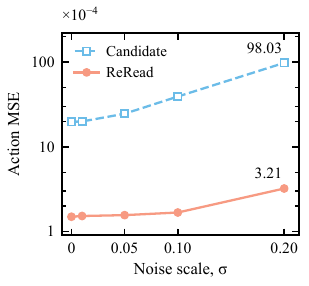}
    \caption{Candidate perturbation.}
    \label{fig:candidate_perturbation}
\end{figure}

When only the caches are changed, swapping in the history from other trajectories of the same task increases the MSE from $1.49\times10^{-4}$ to $18.43\times10^{-4}$ (12.37-fold), close to the level of the native candidate; swapping the temporal order of the two past caches also increases the MSE 3.69-fold, indicating that the corrector uses the temporal order of the caches.
These are the two input paths in Eq.~\eqref{eq:record_sufficient_output}: changes in the candidate are largely canceled by the residual, whereas changes in the caches alter the action through re-reading.
All conditions share the same weights, so the performance differences reflect the effect of cache replacement.

\begin{figure}[!htbp]
    \centering
    \includegraphics[scale=1.15]{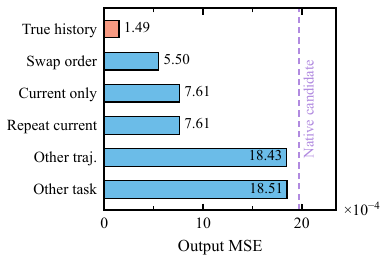}
    \caption{History replacement.}
    \label{fig:history_replacement}
\end{figure}

\begin{figure}[!htbp]
    \centering
    \includegraphics[scale=1.15]{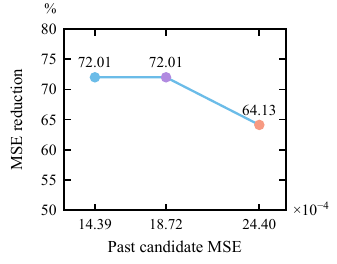}
    \caption{MSE reduction across past prediction error groups.}
    \label{fig:history_error}
\end{figure}

\paragraph{Corrective value of high-error history.}
When a past decision predicts the action inaccurately, is the cache it leaves behind still useful?
We fix the current candidate, the current K/V, and all observations, and rerun the backbone only at the two past anchors with six preset seeds, obtaining six sets of history; in each set, the candidate and the K/V come from the same forward pass.
We divide the six sets into low-, medium-, and high-error groups by the error of the past candidates (protocol in Appendix~\ref*{sec:appendix_open_loop}).
Relative to the current-only readout, the three groups reduce the current MSE by 72.01\%, 72.01\%, and 64.13\%, respectively (Fig.~\ref{fig:history_error}): higher past error yields a slightly smaller gain, but the highest-error group still retains most of the corrective value.

This gain comes from the content of the past K/V rather than from the past states or temporal information. Replacing the past K/V with the current K/V removes the gain; zeroing out only the past K/V while keeping the states and temporal information raises the error to $14.92\times10^{-4}$ (Tab.~\ref*{tab:quality_controls} in the appendix).
With naturally occurring history, the MSE of the past candidates varies 38.45-fold, yet the error reduction after reading the history remains between 77.66\% and 82.04\% (Tab.~\ref*{tab:quality_natural} in the appendix).
\subsection{Is Perception the Missing Piece?}
\label{sec:exp_perception}

\paragraph{External visual encoders.}
If the backbone's failures stem from insufficient perception, adding visual information should yield gains.
We equip the corrector with an external frozen visual branch that reads the four frames $[t-6,t-4,t-2,t]$, and compare four visual encoders (Tab.~\ref{tab:readout_ablations}, group A).
These branches add about 86M pretrained parameters and 86M trainable parameters, yet their average success rates range from 78.91\% to 80.36\%, all below that of ReRead without external vision (81.45\%).
With DINOv3 fixed, the four short-term frames and the current single frame both yield 80.36\%, and removing the cross-decision caches yields 79.09\%.
These results do not support the explanation that the backbone's failures stem from insufficient perception.

\paragraph{Which K/V to read.}
We separately evaluate reading only the action K/V, reading only the visual K/V, and a no-K/V control (Tab.~\ref{tab:readout_ablations}, group B).
Reading only one of the two yields average success rates of 81.09\% and 80.55\%, respectively, both comparable to the 81.45\% obtained by reading the full caches.
The action K/V come from the computation that generates the candidate action; they contain only one third of the tokens in the full caches, yet still suffice for effective correction. The no-K/V control has the same architecture, parameter count, and training setup as ReRead, differing only in that all K/V are removed; its success rate drops to 75.82\%.
\begin{table}[!htbp]
    \centering
    \caption{Closed-loop ablation of readout configurations.}
    \label{tab:readout_ablations}
    \small
    \setlength{\tabcolsep}{2pt}
    \begin{tabular}{@{}lrrr@{}}
        \toprule
        Variant & Clean$\uparrow$ & Random$\uparrow$ & Average$\uparrow$\\
        \midrule
        ReRead & 85.45 & 77.45 & 81.45\\
        \midrule
        \multicolumn{4}{@{}l}{\textit{A. External vision and short-term dense history}}\\
        DINOv3~\cite{dinov3_2025} & 81.82 & 78.91 & 80.36\\
        C-RADIOv3~\cite{radio2025} & 83.27 & 76.36 & 79.82\\
        LingBot-Vision~\cite{lingbotvision2026} & 80.00 & 78.91 & 79.45\\
        V-JEPA 2.1~\cite{vjepa21_2026} & 82.55 & 75.27 & 78.91\\
        \shortstack[l]{DINOv3\\(current image only)} & 82.18 & 78.55 & 80.36\\
        \shortstack[l]{DINOv3\\(current cache only)} & 81.45 & 76.73 & 79.09\\
        \midrule
        \multicolumn{4}{@{}l}{\textit{B. K/V sources}}\\
        Action K/V only & 83.64 & 78.55 & 81.09\\
        Visual K/V only & 83.64 & 77.45 & 80.55\\
        No K/V & 78.18 & 73.45 & 75.82\\
        \midrule
        \multicolumn{4}{@{}l}{\textit{C. History capacity}}\\
        Memory0 & 78.55 & 78.55 & 78.55\\
        Memory1 & 81.09 & 76.36 & 78.73\\
        Memory4 & 81.45 & 75.27 & 78.36\\
        \bottomrule
    \end{tabular}
\end{table}

\subsection{Where Is the Phase Information?}
\label{sec:exp_temporal}
\paragraph{Cross-decision caches.}
When the current frame cannot distinguish phases, past caches can help tell them apart (Eq.~\eqref{eq:history_information_gain}).
In closed-loop evaluation, we change the number of readable past caches from the default 2 to 0, 1, and 4 (Memory0/1/4; Tab.~\ref{tab:readout_ablations}, group C).
Reading only the current cache (Memory0) raises success from 73.45\% to 78.55\%. Adding two past caches raises it further to 81.45\%.
Hard11 is selected by backbone success rate and does not target phase ambiguity, so we further test on a task with clear phase ambiguity below.

\begin{figure*}[!tbp]
    \centering
    \includegraphics[width=\textwidth]{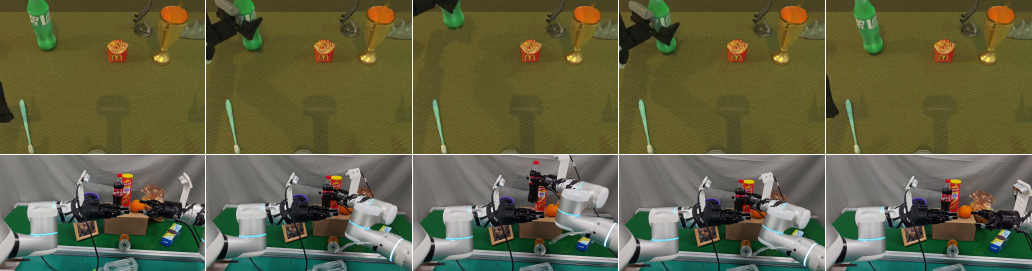}
    \caption{\textbf{Phase-ambiguous task.} Top: simulation. Bottom: real Flexiv dual-arm robot.}
    \label{fig:bottle_task}
\end{figure*}
\paragraph{Phase-ambiguous task.}
Based on RoboTwin 2.0, we build a phase-ambiguous task (Fig.~\ref{fig:bottle_task}, top): pick up a bottle, put it back in place, then open the gripper and return to the initial pose.
The frames before grasping and after placing look similar, yet the next actions differ, so the current frame alone cannot determine the phase.
Fast-WAM is first fine-tuned on 200 expert demonstrations and then frozen. ReRead keeps this backbone frozen and continues training from the corrector weights trained on all 50 RoboTwin 2.0 tasks. The two are compared scene by scene on 50 paired scenes. In simulation, ReRead raises the success rate from 2\% to 50\%. We also set up the same task on a Flexiv dual-arm robot platform, where ReRead raises the real-robot success rate from 0\% to 36\% (Tab.~\ref{tab:bottle_pilot}).
See Appendix~\ref*{sec:appendix_bottle_pilot} for task details.

\begin{table}[!htbp]
    \centering
    \caption{Results on the phase-ambiguous task in simulation and on a real robot.}
    \label{tab:bottle_pilot}
    \small
    \setlength{\tabcolsep}{2pt}
    \begin{tabular}{@{}lrrr@{}}
        \toprule
        Model & Clean$\uparrow$ & Random$\uparrow$ & Average$\uparrow$\\
        \midrule
        \multicolumn{4}{@{}l}{\textit{Simulation}}\\
        Fast-WAM & 4.00 & 0.00 & 2.00\\
        ReRead & \textbf{48.00} & \textbf{52.00} & \textbf{50.00}\\
        \midrule
        \multicolumn{4}{@{}l}{\textit{Real robot}}\\
        Fast-WAM & 0.00 & 0.00 & 0.00\\
        ReRead & \textbf{36.00} & \textbf{36.00} & \textbf{36.00}\\
        \bottomrule
    \end{tabular}
\end{table}

\paragraph{Targeted intervention on phase history.}
\looseness=-1
To test whether ReRead uses phase information in past caches, we intervene at 14 pre-grasp and post-place decisions in 7 scenes.
We fix the current input, the candidate action, and the current K/V, and replace only the caches of the two previous decisions with history from the same or the opposite phase; the replacements come from other trajectories and from the other phase of the same scene.
We measure the change in the corrected action with a directional coordinate: the expert action of the current phase is 0 and the expert reference action of the other phase is 1, so a positive shift means that the corrected action turns toward the other phase (definition in Appendix~\ref*{sec:appendix_phase_intervention}).

Compared with swapping in same-phase history, swapping in opposite-phase history turns the corrected action toward the other phase.
Under cross-trajectory replacement, the action MSE is $127.52\times10^{-4}$ with same-phase history and $521.52\times10^{-4}$ with opposite-phase history; the mean directional shift (opposite phase minus same phase) is $+0.33$, positive in all seven scenes. With the other-phase history from the same scene, the mean shift is $+0.31$.

\begin{figure}[!htbp]
    \centering
    \includegraphics[width=\linewidth,trim=0 7bp 0 0,clip]{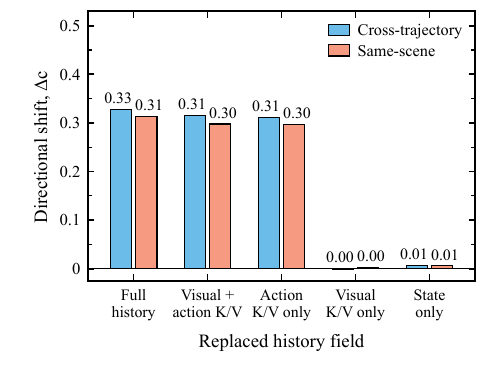}
    \caption{Directional shift after replacing history fields.}
    \label{fig:bottle_phase_paths}
\end{figure}

\looseness=-1
Broken down by replaced field (Fig.~\ref{fig:bottle_phase_paths}), the effect comes mainly from the past action K/V: replacing only the action K/V yields shifts of $+0.31$ and $+0.30$ for the two history sources, close to the $+0.33$ and $+0.31$ obtained by swapping in the full history; replacing only the visual K/V or the past states yields shifts near zero.
Phase information is thus stored in past native caches and read out mainly through action K/V. Past-state replacement does not reproduce the directional shift caused by action-K/V replacement.

\section{Conclusion}
\label{sec:conclusion}
We argue that wrong predictions do not imply wrong representations: a policy can retain the required information yet fail to read it out as the correct action.
To test this view, ReRead re-reads the frozen backbone's native K/V to predict an action residual, without seeing images.
It improves average success rates across three backbones on Hard11 and for Fast-WAM on all 50 tasks.
Interventions and ablations show that correction relies on the backbone's own caches, with phase information read out mainly through past action K/V.
On a Flexiv dual-arm robot, ReRead raises success on the phase-ambiguous task from 0\% to 36\%.
These findings support evaluating prediction and representation separately to understand policy failures. Our training and inference code and model weights are publicly available at \url{https://github.com/tan2355/reread}.

\clearpage
\appendix
\twocolumn[{\centering\Large\bfseries Appendix\par\vspace{12pt}}]
\setcounter{table}{0}
\renewcommand{\thetable}{A\arabic{table}}
\renewcommand{\theHtable}{appendix.\arabic{table}}
\setcounter{figure}{0}
\renewcommand{\thefigure}{A\arabic{figure}}
\renewcommand{\theHfigure}{appendix.\arabic{figure}}
\FloatBarrier
\Needspace{6\baselineskip}
\section{Implementation and Training Details}
\label{sec:appendix_impl}

\paragraph{Model and training.}
ReRead uses a five-layer corrector to read the native K/V. All history-capacity variants use the same candidate action interface and loss function, differing only in the number of readable past caches. Memory1 and Memory4 reuse the continuous temporal metadata encoding; Memory0 keeps the three-slot interface, masks out the past K/V and states before content projection, and excludes past tokens through the attention mask. Table~\ref{tab:impl} lists the Fast-WAM configuration used for the history-capacity ablations. Seen/unseen refers to the task-instruction split.

\begin{table*}[!tbp]
\centering
\caption{History-capacity ablation settings.}
\label{tab:impl}
\small
\setlength{\tabcolsep}{10pt}
\begin{tabular}{@{}p{0.30\textwidth}p{\dimexpr0.70\textwidth-20pt\relax}@{}}
\toprule
Item & Configuration\\
\midrule
Backbone / optimization scope & Frozen Fast-WAM / corrector only\\
Action interface & 20 denoising steps, replan after 20 executed steps, 32×14 action chunk\\
K/V source layers (0-based) & $4,14,24,29$, visual prefill and last action denoising call\\
Corrector & Width 1536, 5 layers, 24 heads, FFN 6144, ReLU\\
Trainable parameters & 198,523,406\\
Training budget & 16 GPUs, batch size 64 per GPU and 1024 globally, 20K steps\\
Optimizer / learning rate & AdamW / peak $10^{-4}$, 1000 warmup steps\\
Training loss & MSE over the full valid action chunk\\
Candidate perturbation & Action-position selection probability 0.2, normalized joint noise std.\ 0.05\\
Training tasks & All 50 RoboTwin 2.0 tasks\\
Data composition & clean:random = 1:10; clean/seen, random/seen+unseen\\
\bottomrule
\end{tabular}
\end{table*}

\paragraph{Cache reading.}
For Fast-WAM, the native K/V are extracted from backbone layers 4, 14, 24, and 29, after normalization and before RoPE. At each source layer, we apply adaptive average pooling to the visual K/V, using a $4\times8$ grid for the head camera and a $4\times4$ grid for each wrist camera. This yields 64 visual tokens per layer, with all 32 action tokens retained. The cache slots store the older past cache, the more recent past cache, and the current cache, in that order. During training, we mask both past caches with probability 0.05 and only the older past cache with probability 0.0475. The current cache remains visible, and the same mask is shared across all source and reader layers.

\subsection{Training Cost and Inference Overhead}
\label{sec:appendix_resource_cost}

We measure ReRead's inference cost on a single A100 80GB GPU. At each decision, the backbone generates a full action chunk of length $H$ using $D$ denoising steps, and ReRead corrects it. Replanning follows after $S$ steps are executed.

\begin{table}[!htbp]
\centering
\caption{End-to-end inference cost.}
\label{tab:inference_cost}
\small
\setlength{\tabcolsep}{5pt}
\begin{tabular}{@{}lrr@{}}
\toprule
Configuration & \shortstack[r]{Mean latency\\(ms)} & \shortstack[r]{Throughput\\(decisions/s)}\\
\midrule
Fast-WAM & 279.89 & 3.573\\
ReRead & 291.55 & 3.430\\
\bottomrule
\end{tabular}
\end{table}

As shown in Tab.~\ref{tab:inference_cost}, adding ReRead to Fast-WAM increases the mean latency per decision by 11.66 ms (4.17\%). Table~\ref{tab:corrector_cost} reports the memory and computation overhead, including temporary memory for packing current and past K/V.

\begin{table}[!htbp]
\centering
\caption{Memory and computation overhead.}
\label{tab:corrector_cost}
\small
\setlength{\tabcolsep}{6pt}
\begin{tabular}{@{}p{0.66\columnwidth}r@{}}
\toprule
Item & Value\\
\midrule
Resident past K/V memory & 9 MiB\\
Temporary K/V packing memory & 13.5 MiB\\
Compute per decision & 87.84 GFLOPs\\
\bottomrule
\end{tabular}
\end{table}

The corrector is trained on $16\times$ H800 GPUs, with per-GPU and global batch sizes of 64 and 1024, respectively. Training runs for 20K steps and takes about 5.80 hours, or 92.73 GPU-hours.

\Needspace{6\baselineskip}
\section{Representation Error, Readout Gap, and Historical Information}
\label{sec:appendix_theory}

\subsection{Projection Decomposition and Error Bounds}
\label{sec:appendix_theory_projection}
\label{sec:appendix_theory_certificate}

We fix the backbone, the trained corrector, and the evaluation distribution. Let $Z=(A^0,\mathcal M)$ and $\Delta=A^\star-A^0$. We assume that the actions and readouts are square-integrable, with the squared norm averaged over action dimensions.
For brevity, we write $\mathcal K(Z)$, $G(Z)$, and $b_\phi(Z)$ from the main text as $\mathcal K$, $G$, and $b_\phi$. Define
$m_Z=\mathbb E[A^\star\mid Z]$ and $r^\star=m_Z-A^0$, and write
\begin{equation}
\begin{aligned}
    \mathcal K
      &=\inf_f\mathbb E\|A^\star-f(Z)\|^2\\
      &=\mathbb E\|A^\star-m_Z\|^2,\\
    \mathcal R(r)
      &=\mathbb E\|A^\star-A^0-r(Z)\|^2\\
      &=\mathcal K+\mathbb E\|r-r^\star\|^2.
\end{aligned}
\label{eq:appendix_projection}
\end{equation}

To prove this decomposition, expand
$A^\star-A^0-r=(A^\star-m_Z)+(r^\star-r)$.
Since $\mathbb E[A^\star-m_Z\mid Z]=0$, the cross term has zero expectation.
Writing $G=\mathbb E\|m_Z-A^0\|^2$ and setting $r=0$ gives
$\mathcal R(0)=\mathcal K+G$.
Since $A^0$ is part of $Z$, any action readout $f(Z)$ can be written as
$A^0+[f(Z)-A^0]$. Thus, residual readout and direct action prediction have the same optimal risk
$\mathcal K$.

Let the approximation error of the trained corrector be
$\varepsilon_\phi=\mathbb E\|r_\phi-r^\star\|^2$.
The same decomposition gives
\begin{equation}
\begin{aligned}
    \mathcal R(r_\phi)&=\mathcal K+\varepsilon_\phi,\\
    \mathcal R(0)-\mathcal R(r_\phi)&=G-\varepsilon_\phi.
\end{aligned}
\label{eq:learned_correction_gain}
\end{equation}
When $\mathcal R(0)>0$, define
\[
b_\phi
=1-\frac{\mathcal R(r_\phi)}{\mathcal R(0)}.
\]
Since $\varepsilon_\phi\geq0$, we have
\[
b_\phi\leq\frac{G}{\mathcal R(0)},
\qquad
\frac{\mathcal K}{\mathcal R(0)}\leq1-b_\phi.
\]
Thus, the achieved error reduction cannot exceed the readout gap and can be used to estimate a lower bound on it;
the corrected risk is no smaller than the representation error, so $1-b_\phi$ upper-bounds the ratio of the representation error to the initial error.
These results concern population risk; in the experiments, we estimate $b_\phi$ using held-out data.

\subsection{Current Caches and the Gain from History}
\label{sec:appendix_theory_history}

Let $Z_c=(A^0,\mathcal C)$ and $Z_m=(Z_c,\mathcal H^-)$, with corresponding conditional means
$m_c,m_m$. Write
\[
\begin{aligned}
K_c&=\mathcal K(Z_c)=\mathbb E\|A^\star-m_c\|^2,\\
K_m&=\mathcal K(Z_m)=\mathbb E\|A^\star-m_m\|^2,
\end{aligned}
\]
and similarly define $G_c=G(Z_c)=\mathbb E\|m_c-A^0\|^2$ and
$G_m=G(Z_m)=\mathbb E\|m_m-A^0\|^2$. By the law of iterated expectations,
$\mathbb E[m_m\mid Z_c]=m_c$. Expanding
$A^\star-m_c=(A^\star-m_m)+(m_m-m_c)$
and eliminating the cross term gives
\begin{equation}
\begin{aligned}
    G_{\rm hist}&=K_c-K_m\\
    &=\mathbb E\|m_m-m_c\|^2=G_m-G_c.
\end{aligned}
    \label{eq:appendix_nested_gain}
\end{equation}
This quantity is the reduction in optimal prediction error after adding history.

Let $\mathcal R_c,\mathcal R_m$ be the actual readout risks under the two inputs, with approximation errors
$\varepsilon_c,\varepsilon_m$. Then
\begin{equation}
    \mathcal R_c-\mathcal R_m
    =G_{\rm hist}+\varepsilon_c-\varepsilon_m.
    \label{eq:actual_vs_optimal_information_gain}
\end{equation}
If the error reduction ratios of both readouts use the same $\mathcal R(0)$ as the baseline, then
\begin{equation}
    b_m-b_c
    =\frac{G_{\rm hist}+\varepsilon_c-\varepsilon_m}
           {\mathcal R(0)}.
    \label{eq:certificate_history_difference}
\end{equation}

\subsection{Bidirectional Interventions on the Candidate and Caches}
\label{sec:appendix_theory_corruption}

Replace the candidate action with a perturbed version
$U=A^0+\eta$, while keeping the caches $\mathcal M$ unchanged.
Under squared loss, the optimal residual and corrected action are
\begin{equation}
\begin{aligned}
    r_\eta^\star(U,\mathcal M)
      &=\mathbb E[A^\star\mid U,\mathcal M]-U,\\
    f_\eta^\star(U,\mathcal M)
      &=\mathbb E[A^\star\mid U,\mathcal M].
\end{aligned}
    \label{eq:appendix_corruption_optimum}
\end{equation}
If, given the caches, the candidate provides no additional information about the target action, i.e.,
$\mathbb E[A^\star\mid U,\mathcal M]=h(\mathcal M)$,
then the optimal residual exactly cancels changes in $U$, and the corrected action is determined solely by the caches.

For a fixed network
$f_\phi(U,\mathcal M)=U+R_\phi(U,\mathcal M)$,
the responses along the two input paths are
\begin{equation}
\begin{aligned}
    d_U
      &=\eta+R_\phi(A^0+\eta,\mathcal M)\\
      &\quad-R_\phi(A^0,\mathcal M),\\
    d_{\mathcal M}
      &=R_\phi(A^0,\mathcal M')
        -R_\phi(A^0,\mathcal M).
\end{aligned}
\label{eq:bidirectional_responses}
\end{equation}
Thus, the response along the candidate input path is determined by both the direct addition term and the residual; the cache input path affects the corrected action only through the residual.
Appendix~\ref{sec:appendix_gaussian} measures the ratio of the RMS of $d_U$ to that of the input perturbation;
Appendices~\ref{sec:appendix_history_intervention} and
\ref{sec:appendix_phase_intervention} measure the action error after cache replacement
and the directional shift, respectively.

\subsection{Distinguishing Phases Despite Past Prediction Errors}
\label{sec:appendix_theory_example}

Fix the current input $Z_c=z$. Let
$p(z)=\Pr(J=+\mid Z_c=z)$, with phase targets $a_-,a_+$,
and let $A^\star=a_J+\epsilon$, where
$\mathbb E[\epsilon\mid Z_c,J,\mathcal H^-]=0$.
Denote the phase probability given the history by
$p_H=\Pr(J=+\mid Z_c=z,\mathcal H^-)$, so that
\begin{equation}
\begin{aligned}
    m_c&=a_-+p(z)(a_+-a_-),\\
    m_m&=a_-+p_H(a_+-a_-).
\end{aligned}
\end{equation}
Define the conditional gain from history as $G_{\rm hist}(z)=\mathbb E[\|m_m-m_c\|^2\mid Z_c=z]$.
The population gain in Eq.~\eqref{eq:appendix_nested_gain} then satisfies $G_{\rm hist}=\mathbb E[G_{\rm hist}(Z_c)]$.
Substituting the conditional means above gives
\begin{equation}
\begin{aligned}
    G_{\rm hist}(z)
    &=\mathbb E[(p_H-p(z))^2\mid Z_c=z]\\
    &\quad\cdot\|a_+-a_-\|^2.
\end{aligned}
    \label{eq:phase_ambiguity_gain}
\end{equation}
The error reduction is positive when the history changes the phase probability and the two phases have different targets.
If the history fully distinguishes the phases, it eliminates the entire error due to phase ambiguity,
$p(z)(1-p(z))\|a_+-a_-\|^2$.

Consider the following example: $J$ takes the values $-1$ and $+1$ with equal probability;
both phases have the same current observation and state, with $F_n=0$ and $A_n^\star=J$;
the past feature and expert action satisfy $F_{n-1}=A_{n-1}^\star=J$.
For the same frozen readout $a(f)=-f$, we have
\begin{equation}
\begin{gathered}
    A_{n-1}^0=-J,\qquad
    A_n^0=0,\\
    (A_{n-1}^0-A_{n-1}^\star)^2=4.
\end{gathered}
\end{equation}
The past prediction is wrong in both phases, but the past feature fully retains $J$. Thus,
\begin{equation}
\begin{gathered}
    r_c^\star=0,\quad
    r_m^\star=J,\\
    K_c=1,\quad
    K_m=0,\quad
    G_{\rm hist}=1.
\end{gathered}
\end{equation}
In this example, the error lies in the action readout, while the information needed to distinguish the phases remains fully preserved in the past caches.

\subsection{Candidate Sampling Variance Decomposition}
\label{sec:appendix_sampling_decomposition}

Let $X$ be the current input and $A^0=\pi_\theta(X;\xi)$,
where the sampling noise $\xi$ is independent of $(X,A^\star)$. Write
$\mu_0=\mathbb E[A^0\mid X]$,
$\mu_\star=\mathbb E[A^\star\mid X]$. Expanding
\[
\begin{aligned}
A^0-A^\star
&=(A^0-\mu_0)+(\mu_0-\mu_\star)\\
&\quad+(\mu_\star-A^\star),
\end{aligned}
\]
and eliminating the cross terms using conditional independence and zero conditional means gives
\begin{equation}
\begin{aligned}
    \mathcal R(0)&=V_0+B_X^2+K_X,\\
    V_0&=\mathbb E\|A^0-\mu_0\|^2,\\
    B_X^2&=\mathbb E\|\mu_0-\mu_\star\|^2,\\
    K_X&=\mathbb E\|A^\star-\mu_\star\|^2.
\end{aligned}
\label{eq:sampling_risk_decomposition}
\end{equation}
The three terms are the average conditional variance of the candidate action, the mean squared error between the conditional means of the candidate and target actions,
and the average conditional variance of the target action. Combining Eq.~\eqref{eq:sampling_risk_decomposition} with $\mathcal R(0)=\mathcal K+G$ gives
\[
G=V_0+B_X^2+K_X-\mathcal K.
\]
Appendix~\ref{sec:appendix_sampling_measurement} uses six fixed candidate sampling seeds
to estimate $V_0$ and directly measures the error reduction from averaging the candidates.

\subsection{Valid Action Elements and Held-out Measurement}
\label{sec:appendix_theory_mask}

Training and evaluation use nonnegative weights $w_k\geq0$ to select valid action elements:
training uses the label mask, while open-loop evaluation uses an evaluation mask over the first $S=20$ steps.
Assume $0<\mu=\mathbb E\sum_kw_k<\infty$ and finite weighted second moments. Then
\begin{equation}
\begin{aligned}
    \mathcal R_w(r)
      &=\frac{\mathbb E\sum_kw_k(\Delta_k-r_k(Z))^2}{\mu},\\
    r_{w,k}^\star(Z)
      &=\frac{\mathbb E[w_k\Delta_k\mid Z]}
             {\mathbb E[w_k\mid Z]}.
\end{aligned}
    \label{eq:appendix_weighted_readout}
\end{equation}
When the denominator is zero, set $r_{w,k}^\star=0$. Weighted orthogonality,
$\mathbb E[w_k(\Delta_k-r_{w,k}^\star)\mid Z]=0$, gives
\begin{equation}
\begin{aligned}
    \mathcal R_w(r)
      &=\mathcal K_w
       +\frac{\mathbb E\sum_kw_k(r_k-r_{w,k}^\star)^2}{\mu},\\
    \mathcal K_w&=\mathcal R_w(r_w^\star).
\end{aligned}
    \label{eq:appendix_weighted_decomposition}
\end{equation}
Thus, the error decomposition and readout bounds in the main text also apply to weighted risk over valid action elements.
With the corrector fixed, we compute the following over decisions $i$ on the held-out trajectories:
\begin{equation}
\begin{aligned}
    \widehat{\mathcal R}(r)
      &=\frac{\sum_{i,k}w_{ik}
          (A^\star_{ik}-A^0_{ik}-r_k(Z_i))^2}
         {\sum_{i,k}w_{ik}},\\
    \widehat b_\phi
      &=1-\frac{\widehat{\mathcal R}(r_\phi)}
              {\widehat{\mathcal R}(0)},
\end{aligned}
    \label{eq:empirical_readout_certificate}
\end{equation}
where $\widehat{\mathcal R}(0)>0$. Appendix~\ref{sec:appendix_readout_gap} provides the measurement settings.

\Needspace{6\baselineskip}
\section{Open-loop Analysis and History Interventions}
\label{sec:appendix_open_loop}

\subsection{Readout Gap and Sampling Variance}
\label{sec:appendix_readout_gap}

With ReRead's weights fixed, we compare three conditions on the 22 held-out expert trajectories of Hard11: the native candidate, reading only the current cache, and reading the current and two previous caches. Errors are aggregated over valid action elements using the weights in Eq.~\eqref{eq:empirical_readout_certificate}.

\paragraph{Sampling variance.}
\label{sec:appendix_sampling_measurement}
On the same trajectories, we generate candidate actions with six nonzero denoising seeds, estimate the sampling variance, and compute the MSE of the mean of the six candidates. Table~\ref{tab:sampling_variance} reports the results.

\begin{table}[!htbp]
\centering
\caption{Candidate error and sampling variance ($10^{-4}$).}
\label{tab:sampling_variance}
\small
\setlength{\tabcolsep}{9pt}
\begin{tabular}{@{}lr@{}}
\toprule
Statistic & Value\\
\midrule
Mean single-candidate MSE & 20.48\\
MSE of the six-candidate mean & 19.76\\
Sampling variance & 0.86\\
Risk minus sampling variance & 19.62\\
\bottomrule
\end{tabular}
\end{table}

\subsection{History Replacement}
\label{sec:appendix_history_intervention}

We evaluate ReRead in open loop on the 22 held-out trajectories of Hard11. Figure~5 in the main text aggregates the clean/seen and random/seen results over valid action elements using Eq.~\eqref{eq:empirical_readout_certificate}; Tab.~\ref{tab:history_intervention_cases} reports the MSE for each setting and adds the random/unseen results under the same protocol. Here, seen/unseen refers to the task-instruction split. All interventions keep the candidate action, current images, robot state, target instruction, initial denoising noise, and current K/V unchanged, and take effect only for the current cache read.

\paragraph{History conditions.}
The cache slots store the older past cache, the more recent past cache, and the current cache, in that order. Each slot contains visual K/V, action K/V, and the robot state. We compare six history conditions:
\begin{enumerate}
    \item \textbf{True history:} read the caches of the current and two previous decisions.
    \item \textbf{Current only:} mask out the K/V and states of both past caches, keeping the current cache.
    \item \textbf{Repeat current:} copy the current K/V and state into the past slots, preserving each slot's age encoding and number of visible tokens.
    \item \textbf{Swap order:} swap the contents of the two past caches, preserving each slot's age encoding.
    \item \textbf{Other trajectories of the same task:} replace the past caches with those of two consecutive decisions from another trajectory of the same task.
    \item \textbf{Other task:} replace the past caches with those of two consecutive decisions from another task.
\end{enumerate}
All content replacements apply to K, V, and the state together.

\paragraph{Selecting replacement trajectories.}
Replacement caches come from other training trajectories. We select three distinct replacement trajectories per task and clean/random setting, 66 trajectories in total; the replacements contain at least four consecutive decision anchors spaced $S$ steps apart. The replacement and target trajectories use the same clean/random setting, and their past anchors are matched by relative decision progress.

\begin{table*}[!tbp]
\centering
\caption{History replacement MSE ($10^{-4}$).}
\label{tab:history_intervention_cases}
\small
\setlength{\tabcolsep}{10pt}
\begin{tabular}{@{}lrrr@{}}
\toprule
History condition & Clean/seen & Random/seen & Random/unseen\\
\midrule
Native candidate & 26.561 & 12.867 & 13.219\\
True history & 1.506 & 1.477 & 1.514\\
Current only & 8.990 & 6.224 & 6.195\\
Repeat current & 8.995 & 6.223 & 6.198\\
Swap order & 5.618 & 5.387 & 5.347\\
Other trajectory (same task) & 22.629 & 14.201 & 14.328\\
Other task & 24.901 & 12.072 & 12.411\\
\bottomrule
\end{tabular}
\end{table*}

\subsection{Response to Candidate Perturbations}
\label{sec:appendix_gaussian}

We use the same ReRead weights, 22 held-out trajectories, and candidate perturbation settings as in the main text. All noise levels share the same random mask and Gaussian samples, with only the perturbation scale changed. We measure perturbation transfer by the RMS ratio of the change in the corrected action to the input perturbation, $\mathrm{RMS}(d_U)/\mathrm{RMS}(\eta)$, where $d_U$ is defined in Eq.~\eqref{eq:bidirectional_responses}. At $\sigma=0.05$ and $0.20$, the ratios are 12.3\% and 15.1\%, respectively.

\subsection{Past Prediction Error and Corrective Value}
\label{sec:appendix_history_quality}

We pair each past candidate's prediction error with the cache generated in the same forward pass to test whether the cache remains useful for correction when the past prediction is wrong. The experiments fix ReRead's weights and use the 22 held-out trajectories of Hard11. For decision $w$, let $Q_w$ be the mean error of the native candidates at the two previous decisions. We fix the current observations, candidate action, and current K/V, and use six preset denoising seeds only on the past observations to generate six sets of history. We divide them into low-, medium-, and high-error groups by $Q_w$.

Table~\ref{tab:quality_controls} reports the MSE for each error group and the controls on history content. Here, K/V includes both visual and action K/V. The zero-K/V and current-K/V controls retain past states and temporal information; the three error groups use native history.

\begin{table*}[!tbp]
\centering
\caption{History content controls (MSE in $10^{-4}$).}
\label{tab:quality_controls}
\small
\setlength{\tabcolsep}{8pt}
\begin{tabular}{@{}p{0.52\textwidth}rr@{}}
\toprule
History condition & \shortstack[r]{Past candidate\\MSE} & \shortstack[r]{Current action\\MSE}\\
\midrule
Mask past K/V and states & --- & 7.61\\
Zero past K/V & --- & 14.92\\
Current K/V in past slots & --- & 7.81\\
Low-error history K/V & 14.39 & 2.13\\
Medium-error history K/V & 18.72 & 2.13\\
High-error history K/V & 24.40 & 2.73\\
\bottomrule
\end{tabular}
\end{table*}

For naturally occurring history, we compute the quantiles of $Q_w$ at $1/3$ and $2/3$ separately for each task and clean/random setting. These thresholds divide the decisions into low-, medium-, and high-error history groups, with results in Tab.~\ref{tab:quality_natural}.

\begin{table*}[!tbp]
\centering
\caption{History error groups (MSE in $10^{-4}$).}
\label{tab:quality_natural}
\small
\setlength{\tabcolsep}{9pt}
\begin{tabular}{@{}p{0.30\textwidth}rrrr@{}}
\toprule
History error group & \shortstack[r]{Past candidate\\MSE} & Current only & True history & \shortstack[r]{MSE reduction\\(\%)}\\
\midrule
Low error & 1.18 & 9.73 & 1.81 & 81.40\\
Medium error & 5.07 & 7.43 & 1.66 & 77.66\\
High error & 45.37 & 5.79 & 1.04 & 82.04\\
Both past candidates: high error & 44.62 & 4.05 & 0.83 & 79.51\\
\bottomrule
\end{tabular}
\end{table*}

\Needspace{6\baselineskip}
\section{Closed-loop Results and Ablations}
\label{sec:appendix_closed_loop}

\subsection{Three Backbones on Hard11}
\label{sec:appendix_local_backbones}

\begin{table*}[!tbp]
\centering
\caption{Successful episodes on Hard11. C/R: clean/random.}
\label{tab:local_backbone_tasks}
\small
\setlength{\tabcolsep}{3.2pt}
\renewcommand{\arraystretch}{1.0}
\begin{tabular}{@{}l*{12}{r}@{}}
\toprule
& \multicolumn{4}{c}{Fast-WAM} & \multicolumn{4}{c}{$\pi_{0.5}$} & \multicolumn{4}{c}{LingBot-VLA}\\
\cmidrule(lr){2-5}\cmidrule(lr){6-9}\cmidrule(lr){10-13}
& \multicolumn{2}{c}{Backbone only} & \multicolumn{2}{c}{ReRead}
& \multicolumn{2}{c}{Backbone only} & \multicolumn{2}{c}{ReRead}
& \multicolumn{2}{c}{Backbone only} & \multicolumn{2}{c}{ReRead}\\
\cmidrule(lr){2-3}\cmidrule(lr){4-5}\cmidrule(lr){6-7}\cmidrule(lr){8-9}\cmidrule(lr){10-11}\cmidrule(lr){12-13}
Task & C & R & C & R & C & R & C & R & C & R & C & R\\
\midrule
\texttt{open\_microwave} & 13 & 8 & 25 & 23 & 13 & 11 & 5 & 6 & 1 & 1 & 21 & 23\\
\texttt{hanging\_mug} & 10 & 19 & 16 & 10 & 0 & 0 & 5 & 1 & 5 & 6 & 10 & 9\\
\texttt{turn\_switch} & 19 & 18 & 16 & 15 & 6 & 5 & 13 & 11 & 10 & 16 & 16 & 17\\
\texttt{place\_can\_basket} & 20 & 16 & 19 & 18 & 14 & 8 & 17 & 17 & 10 & 11 & 17 & 15\\
\texttt{move\_stapler\_pad} & 20 & 15 & 20 & 17 & 5 & 4 & 10 & 5 & 13 & 19 & 19 & 18\\
\texttt{stack\_bowls\_three} & 20 & 21 & 21 & 24 & 9 & 10 & 13 & 14 & 14 & 15 & 19 & 19\\
\texttt{pick\_diverse\_bottles} & 23 & 19 & 23 & 20 & 5 & 13 & 14 & 7 & 17 & 17 & 24 & 19\\
\texttt{place\_mouse\_pad} & 21 & 20 & 23 & 20 & 2 & 5 & 10 & 14 & 12 & 15 & 23 & 22\\
\texttt{handover\_block} & 22 & 15 & 24 & 22 & 10 & 5 & 10 & 7 & 10 & 10 & 15 & 15\\
\texttt{place\_object\_basket} & 22 & 20 & 24 & 22 & 9 & 12 & 19 & 20 & 24 & 24 & 23 & 19\\
\texttt{move\_can\_pot} & 21 & 22 & 24 & 22 & 10 & 11 & 9 & 12 & 12 & 13 & 20 & 21\\
\midrule
Average (\%) & \multicolumn{2}{c}{73.45} & \multicolumn{2}{c}{81.45} & \multicolumn{2}{c}{30.36} & \multicolumn{2}{c}{43.45} & \multicolumn{2}{c}{50.00} & \multicolumn{2}{c}{73.45}\\
\bottomrule
\end{tabular}
\end{table*}

\begingroup
\raggedright
\setlength{\parindent}{1em}
\Urlmuskip=0mu\relax
\def\UrlBreakPenalty{100}
\expandafter\def\expandafter\UrlBreaks\expandafter{%
  \UrlBreaks\do\a\do\b\do\c\do\d\do\e\do\f\do\g\do\h\do\i\do\j\do\k\do\l\do\m\do\n\do\o\do\p\do\q\do\r\do\s\do\t\do\u\do\v\do\w\do\x\do\y\do\z%
  \do\A\do\B\do\C\do\D\do\E\do\F\do\G\do\H\do\I\do\J\do\K\do\L\do\M\do\N\do\O\do\P\do\Q\do\R\do\S\do\T\do\U\do\V\do\W\do\X\do\Y\do\Z%
  \do\0\do\1\do\2\do\3\do\4\do\5\do\6\do\7\do\8\do\9}
We use the authors' Fast-WAM Base checkpoint for RoboTwin from \nolinkurl{yuanty/fastwam}, revision \nolinkurl{8eaceeb24c3cc92ff2a9c9a9d266a4941b836705}. The weight file is \nolinkurl{robotwin_uncond_3cam_384.pt}, paired with the normalization statistics in \nolinkurl{robotwin_uncond_3cam_384_dataset_stats.json}.

For $\pi_{0.5}$, we use the RoboTwin 2.0 checkpoint fine-tuned and released by Motus from \nolinkurl{motus-robotics/pi0.5_robotwin2}, revision \nolinkurl{effc6e1a99bd4d0edf9447e8ba313def67649cfe}. Motus converted the official $\pi_{0.5}$ weights from JAX to PyTorch and fine-tuned them on 27,500 trajectories across 50 RoboTwin 2.0 tasks, with 50 clean and 500 random trajectories per task. The action chunk length is 32.

For LingBot-VLA, we use the authors' 4B RoboTwin post-trained checkpoint without depth inputs from \nolinkurl{robbyant/lingbot-vla-4b-posttrain-robotwin}, revision \nolinkurl{fb71a2c9749ccfedbb7290c2c3f0e5e7c7305c9e}.

\par
\endgroup

We keep these backbones frozen and train a separate ReRead corrector from scratch on offline demonstrations from all 50 RoboTwin 2.0 tasks.

ReRead reads native K/V from layers 2, 8, 14, and 17 of $\pi_{0.5}$ and layers 5, 17, 29, and 35 of LingBot-VLA. The candidate action provides the query, and the corrector reads the caches of the current and two previous decisions to predict an action residual.

Each model is evaluated for 25 episodes per task and setting on Hard11, 550 episodes in total. We evaluate all three backbone-only baselines using the inference settings provided with their respective checkpoints. Table~\ref{tab:local_backbone_tasks} reports the per-task results.

\begin{table*}[!tp]
\centering
\caption{Per-task success rates (\%). C/R: clean/random.}
\label{tab:full_tasks}
\scriptsize
\setlength{\tabcolsep}{1.2pt}
\renewcommand{\arraystretch}{1.15}
\begin{tabular*}{\textwidth}{@{\extracolsep{\fill}}l*{18}{r}@{}}
\toprule
& \multicolumn{2}{c}{$\pi_{0.5}$} & \multicolumn{2}{c}{LingBot-VLA} & \multicolumn{2}{c}{Motus} & \multicolumn{2}{c}{HyVLA-0.5} & \multicolumn{2}{c}{GeoMem-VLA} & \multicolumn{2}{c}{LingBot-VA} & \multicolumn{2}{c}{Fast-WAM} & \multicolumn{2}{c}{Fast-WAM$^{\dagger}$} & \multicolumn{2}{c}{ReRead} \\
\cmidrule(lr){2-3} \cmidrule(lr){4-5} \cmidrule(lr){6-7} \cmidrule(lr){8-9} \cmidrule(lr){10-11} \cmidrule(lr){12-13} \cmidrule(lr){14-15} \cmidrule(lr){16-17} \cmidrule(lr){18-19}
Task & C & R & C & R & C & R & C & R & C & R & C & R & C & R & C & R & C & R \\
\midrule
Adjust Bottle & 79 & 83 & \textbf{100} & \textbf{100} & 89 & 93 & 99 & 99 & \textbf{100} & \textbf{100} & 90 & 94 & \textbf{100} & \textbf{100} & \textbf{100} & \textbf{100} & \textbf{100} & \textbf{100} \\
Beat Block Hammer & 63 & 50 & 87 & 91 & 95 & 88 & 99 & 99 & 99 & 94 & 96 & 98 & 99 & 97 & 97 & 95 & \textbf{100} & \textbf{100} \\
Blocks Ranking RGB & 43 & 35 & 92 & 91 & 99 & 97 & 99 & 99 & 99 & 95 & 99 & 98 & \textbf{100} & \textbf{100} & \textbf{100} & \textbf{100} & \textbf{100} & \textbf{100} \\
Blocks Ranking Size & 8 & 14 & 66 & 73 & 75 & 63 & 93 & 94 & 87 & 86 & 94 & 96 & 94 & \textbf{98} & 91 & 91 & \textbf{96} & 94 \\
Click Alarmclock & 97 & 93 & 93 & 26 & \textbf{100} & \textbf{100} & \textbf{100} & \textbf{100} & 92 & 97 & 99 & \textbf{100} & \textbf{100} & \textbf{100} & \textbf{100} & \textbf{100} & \textbf{100} & \textbf{100} \\
Click Bell & 75 & 76 & 32 & 19 & \textbf{100} & \textbf{100} & \textbf{100} & 99 & 97 & 95 & \textbf{100} & \textbf{100} & \textbf{100} & \textbf{100} & \textbf{100} & \textbf{100} & \textbf{100} & \textbf{100} \\
Dump Bin Bigbin & 30 & 42 & 97 & 92 & 95 & 91 & 96 & 98 & 97 & 98 & 89 & 96 & 97 & 96 & 95 & \textbf{99} & \textbf{98} & 96 \\
Grab Roller & 90 & 89 & \textbf{100} & 99 & \textbf{100} & \textbf{100} & \textbf{100} & \textbf{100} & 98 & 98 & \textbf{100} & \textbf{100} & \textbf{100} & \textbf{100} & \textbf{100} & \textbf{100} & \textbf{100} & \textbf{100} \\
Handover Block & 18 & 19 & 80 & 83 & 86 & 73 & 93 & 73 & 93 & \textbf{92} & \textbf{99} & 78 & 95 & 81 & 91 & 78 & 92 & 88 \\
Handover Mic & 28 & 18 & 94 & 98 & 78 & 63 & 87 & 98 & \textbf{100} & 98 & 94 & 96 & 99 & \textbf{100} & \textbf{100} & \textbf{100} & 99 & 98 \\
Hanging Mug & 3 & 3 & 32 & 27 & 38 & 38 & 37 & 33 & 44 & 40 & 40 & 28 & \textbf{58} & 62 & \textbf{58} & \textbf{67} & 47 & 51 \\
Lift Pot & 0 & 0 & \textbf{100} & 99 & 96 & 99 & 99 & 99 & \textbf{100} & \textbf{100} & \textbf{100} & 99 & \textbf{100} & \textbf{100} & \textbf{100} & 99 & \textbf{100} & \textbf{100} \\
Move Can Pot & 29 & 27 & 79 & 84 & 34 & 74 & 94 & \textbf{99} & 87 & 91 & 94 & 97 & 90 & 88 & 88 & 89 & \textbf{96} & 94 \\
Move Pillbottle Pad & 33 & 29 & 93 & 94 & 93 & 96 & 96 & 96 & \textbf{100} & 98 & 99 & \textbf{99} & \textbf{100} & \textbf{99} & 99 & \textbf{99} & \textbf{100} & \textbf{99} \\
Move Playingcard Away & 59 & 67 & 96 & 99 & \textbf{100} & 96 & 98 & 96 & \textbf{100} & 99 & \textbf{100} & 99 & \textbf{100} & \textbf{100} & 99 & 99 & \textbf{100} & \textbf{100} \\
Move Stapler Pad & 16 & 18 & 74 & 49 & 83 & 85 & \textbf{94} & \textbf{95} & 78 & 68 & 91 & 79 & 77 & 64 & 80 & 60 & 80 & 66 \\
Open Laptop & 19 & 35 & 96 & 96 & 95 & 91 & \textbf{99} & 98 & \textbf{99} & \textbf{100} & 92 & 94 & 98 & \textbf{100} & 98 & 99 & 96 & \textbf{100} \\
Open Microwave & 35 & 37 & 91 & 75 & 95 & 91 & 72 & 64 & \textbf{99} & \textbf{99} & 82 & 86 & 62 & 45 & 45 & 45 & 96 & 88 \\
Pick Diverse Bottles & 5 & 3 & 79 & 86 & \textbf{90} & \textbf{91} & 88 & 72 & 73 & 72 & 89 & 82 & 80 & 85 & 88 & 82 & 86 & 82 \\
Pick Dual Bottles & 10 & 6 & 82 & 95 & 96 & 90 & 82 & 82 & 84 & 89 & \textbf{100} & \textbf{99} & \textbf{100} & 96 & 97 & 97 & 97 & 95 \\
Place A2B Left & 62 & 60 & 86 & 83 & 88 & 79 & 77 & 75 & 80 & 82 & 97 & 93 & 95 & 93 & 98 & \textbf{94} & \textbf{99} & \textbf{94} \\
Place A2B Right & 62 & 57 & 74 & 77 & 91 & 87 & 74 & 78 & 75 & 72 & \textbf{97} & 95 & 93 & \textbf{99} & 95 & 94 & \textbf{97} & 96 \\
Place Bread Basket & 48 & 56 & 92 & 93 & 91 & 94 & 95 & 93 & \textbf{97} & 93 & \textbf{97} & 95 & 91 & 93 & 95 & \textbf{96} & 92 & 95 \\
Place Bread Skillet & 38 & 46 & 90 & 89 & 86 & 83 & 94 & 90 & 89 & 89 & \textbf{95} & 90 & 90 & \textbf{93} & 88 & 88 & 94 & 90 \\
Place Burger Fries & 66 & 70 & 95 & 96 & 98 & 98 & \textbf{99} & 98 & 95 & 99 & 97 & 95 & 96 & 99 & 97 & 98 & 96 & \textbf{100} \\
Place Can Basket & 19 & 25 & 68 & 78 & 81 & 76 & \textbf{91} & 81 & 87 & \textbf{93} & 81 & 84 & 71 & 69 & 68 & 70 & 70 & 71 \\
Place Cans Plasticbox & 40 & 47 & 97 & \textbf{100} & 98 & 94 & \textbf{100} & \textbf{100} & 99 & \textbf{100} & \textbf{100} & 99 & 99 & 96 & \textbf{100} & 98 & 98 & 99 \\
Place Container Plate & 71 & 78 & 99 & 99 & 98 & 99 & 98 & 99 & \textbf{100} & \textbf{100} & 99 & 97 & 96 & \textbf{100} & 99 & \textbf{100} & 99 & \textbf{100} \\
Place Dual Shoes & 12 & 7 & 80 & 83 & 93 & 87 & 94 & \textbf{95} & 85 & 81 & 94 & 89 & 94 & 88 & 94 & 93 & \textbf{100} & 90 \\
Place Empty Cup & 75 & 86 & \textbf{100} & \textbf{100} & 99 & 98 & \textbf{100} & \textbf{100} & \textbf{100} & \textbf{100} & \textbf{100} & \textbf{100} & \textbf{100} & \textbf{100} & \textbf{100} & \textbf{100} & \textbf{100} & \textbf{100} \\
Place Fan & 25 & 36 & 91 & 79 & 91 & 87 & 96 & 98 & 98 & \textbf{100} & \textbf{99} & 93 & 96 & 96 & 96 & 92 & 98 & 96 \\
Place Mouse Pad & 21 & 26 & 82 & 78 & 66 & 68 & 88 & 93 & 90 & 92 & \textbf{93} & \textbf{96} & 83 & 89 & 79 & 80 & 86 & 85 \\
Place Object Basket & 43 & 36 & 90 & 91 & 81 & 87 & 86 & 85 & 83 & 86 & 91 & 88 & 89 & 88 & 84 & 79 & \textbf{93} & \textbf{93} \\
Place Object Scale & 40 & 49 & 84 & 90 & 88 & 85 & 90 & 89 & 92 & 92 & \textbf{96} & 95 & 90 & \textbf{97} & 94 & 89 & \textbf{96} & 95 \\
Place Object Stand & 74 & 65 & 97 & 93 & 98 & 97 & 94 & \textbf{98} & 97 & 94 & \textbf{99} & 96 & 90 & 94 & 96 & 96 & 97 & 96 \\
Place Phone Stand & 49 & 53 & 92 & 93 & 87 & 86 & 91 & 94 & 90 & \textbf{100} & \textbf{97} & 97 & \textbf{97} & 99 & 95 & 98 & 92 & 99 \\
Place Shoe & 57 & 61 & \textbf{99} & 94 & \textbf{99} & 97 & 98 & \textbf{100} & \textbf{99} & 97 & 98 & 98 & 96 & 99 & 87 & 96 & 91 & 97 \\
Press Stapler & 80 & 70 & 90 & 88 & 93 & 98 & 88 & 80 & \textbf{100} & \textbf{100} & 85 & 82 & 90 & 97 & 91 & 92 & 93 & 92 \\
Put Bottles Dustbin & 12 & 9 & 88 & 92 & 81 & 79 & 89 & 85 & 84 & 88 & 87 & 91 & \textbf{95} & 90 & 91 & 93 & 90 & \textbf{95} \\
Put Object Cabinet & 24 & 15 & 92 & 86 & 88 & 71 & 78 & 83 & 88 & 85 & 85 & 87 & \textbf{94} & 89 & 91 & 90 & 92 & \textbf{91} \\
Rotate QRcode & 47 & 56 & 93 & 84 & 89 & 73 & 93 & \textbf{98} & 87 & 89 & \textbf{96} & 91 & 93 & 89 & 90 & 81 & 93 & 79 \\
Scan Object & 42 & 38 & 91 & \textbf{97} & 67 & 66 & 91 & 94 & 95 & 95 & \textbf{96} & 91 & 89 & 92 & 91 & 85 & 95 & 84 \\
Shake Bottle & 91 & \textbf{100} & 99 & \textbf{100} & \textbf{100} & 97 & \textbf{100} & 99 & \textbf{100} & \textbf{100} & \textbf{100} & 97 & \textbf{100} & \textbf{100} & \textbf{100} & \textbf{100} & \textbf{100} & 99 \\
Shake Bottle Horizontally & 96 & \textbf{100} & \textbf{100} & \textbf{100} & \textbf{100} & 98 & \textbf{100} & 99 & \textbf{100} & \textbf{100} & \textbf{100} & 99 & \textbf{100} & \textbf{100} & \textbf{100} & \textbf{100} & \textbf{100} & \textbf{100} \\
Stack Blocks Three & 15 & 16 & 92 & \textbf{99} & 91 & 95 & 98 & 97 & 81 & 84 & \textbf{99} & 98 & 95 & 97 & 77 & 74 & 87 & 80 \\
Stack Blocks Two & 48 & 56 & \textbf{100} & \textbf{100} & \textbf{100} & 98 & 98 & 99 & 99 & 99 & \textbf{100} & 98 & \textbf{100} & \textbf{100} & \textbf{100} & \textbf{100} & \textbf{100} & \textbf{100} \\
Stack Bowls Three & 33 & 35 & 72 & 83 & 79 & 87 & 87 & 83 & \textbf{91} & \textbf{91} & 86 & 83 & 80 & 81 & 85 & 86 & 86 & 89 \\
Stack Bowls Two & 78 & 66 & 92 & 95 & 98 & 98 & 98 & 97 & \textbf{99} & \textbf{99} & 94 & 98 & 92 & 98 & 93 & \textbf{99} & 97 & 95 \\
Stamp Seal & 36 & 23 & 76 & 86 & 93 & 92 & 85 & 81 & 86 & 88 & \textbf{96} & \textbf{97} & 90 & 94 & 89 & 92 & 91 & 94 \\
Turn Switch & 5 & 6 & 61 & 65 & \textbf{84} & \textbf{78} & 50 & 49 & 63 & 65 & 44 & 45 & 61 & 59 & 64 & 68 & 55 & 61 \\
\midrule
Average & 42.98 & 43.84 & 86.50 & 85.34 & 88.66 & 87.02 & 90.90 & 90.06 & 91.10 & 91.24 & 92.90 & 91.50 & 91.88 & 91.78 & 91.06 & 90.40 & \textbf{93.20} & \textbf{92.12} \\
\bottomrule
\end{tabular*}
\end{table*}

\Needspace{8\baselineskip}
\subsection{Per-task Results on All 50 Tasks}
\label{sec:appendix_full_tasks}

Table~\ref{tab:full_tasks} summarizes the success rates of 9 models on all 50 tasks of RoboTwin 2.0. All results except ReRead and Fast-WAM$^{\dagger}$ are taken from the corresponding papers. We evaluate ReRead and Fast-WAM$^{\dagger}$ for 100 episodes per task and setting, with Fast-WAM$^{\dagger}$ using the checkpoint specified in Appendix~\ref{sec:appendix_local_backbones}.

\Needspace{4\baselineskip}
\subsection{External Visual Branches}
\label{sec:appendix_visual}

\paragraph{Architecture and visual inputs.}
Standard ReRead reads the frozen backbone's native K/V. To test whether external visual features provide additional information for action correction, we add a frozen visual encoder and a trainable branch that writes a separate correction memory. The native K/V and the backbone's three-camera input remain unchanged. The external branch reads four frames \mbox{$[t-6,t-4,t-2,t]$} from the head camera at $384\times384$ resolution, where $t$ indexes trajectory rows during training and executed actions during deployment. This short-term window is maintained separately from the caches of previous decisions. At the start of an episode, missing frames are filled with the earliest observation and excluded by a validity mask.

\paragraph{Conditioned K/V writer.}
With DINOv3, each frame yields $24\times24$ normalized patch tokens of dimension 768. We retain all 2,304 patch tokens across four frames, excluding CLS and register tokens. Layer normalization and a linear projection map the tokens to width 1,024, followed by learned joint temporal and spatial position embeddings. Queries combine projected, normalized current action K/V, the frozen backbone's state feature, and layer and action-position embeddings. The state feature reuses the backbone's proprioceptive encoder and Action DiT context mapping. The four source layers and 32 action positions per layer give 128 queries. A single three-layer writer processes all queries, using pre-norm cross-attention and feed-forward residual updates with width 1,024, 16 heads, FFN width 4,096, and GELU. The visual tokens with position embeddings supply both keys and values. After output normalization, separate linear projections produce correction keys and values, each of shape $4\times32\times1536$.

\paragraph{Memory and gated reading.}
Both memories store records from the current and two previous decisions, with shared history-validity and dropout masks. Past correction records are detached before caching; the current record remains differentiable. Age encodings are added to the correction keys at each read. After query self-attention, each of the corrector's five layers reads the native and correction memories separately. The two reads share the query projection and use separate key, value, and output projections and softmax normalizations. Using the query notation in Sec.~\ref{sec:reader}, the update is
\begin{equation*}
    q_j''=q_j'+o_j^{\mathrm{native}}
      +\tanh(g_j)\,o_j^{\mathrm{corr}},
\end{equation*}
where $o_j^{\mathrm{native}}$ and $o_j^{\mathrm{corr}}$ are the two attention outputs, and $g_j$ is a scalar learned independently in each layer and initialized to zero. The feed-forward update follows, and the final output head predicts the action residual. With the backbone and visual encoder frozen, we train the writer, readers, gates, and output head using only the final action MSE.

\paragraph{Visual encoder variants.}
We compare DINOv3 (ViT-B/16), LingBot-Vision (distilled Base version), V-JEPA 2.1 (ViT-B/16), and C-RADIO (C-RADIOv3-B). All variants use the same causal frame sampling and correction memory interface. DINOv3, LingBot-Vision, and C-RADIO retain four per-frame feature slices; V-JEPA 2.1 uses a temporal tubelet size of 2 to produce two temporal slices. The four-frame DINOv3 variant adds 85,842,949 trainable parameters (85.84M) to standard ReRead.

\subsection{K/V Sources and the No-K/V Control}
\label{sec:appendix_kv_source}

The three configurations---reading both visual and action K/V, reading only action K/V, and reading only visual K/V---are trained independently and evaluated for 25 episodes per task and setting. The no-K/V control retains the candidate action, the current and past robot states, and the executed prefix.

\Needspace{6\baselineskip}
\section{Phase-ambiguous Task and Real-robot Experiments}
\label{sec:appendix_bottle_pilot}

\subsection{Closed-loop Evaluation}
\paragraph{Simulation.}
The training set contains 200 successful expert demonstrations, with 20\% clean and 80\% random. We first fine-tune the released Fast-WAM RoboTwin checkpoint for 2K steps with its native video and action objectives, then freeze the backbone and continue training ReRead for 2K steps from the corrector weights trained on all 50 RoboTwin 2.0 tasks using only MSE. Both stages use a global batch size of 8 and a learning rate of $10^{-5}$. The instruction does not explicitly identify the current phase.

We evaluate on 50 paired scenes, with 25 scenes in each of the clean and random settings. Each model runs once in each scene, giving 100 episodes in total. Both use the same initial observations, instruction, and candidate denoising seed, with $D$ denoising steps and replanning after $S$ executed steps. Success requires completing the grasp, placement, gripper opening, and return to the initial pose. The episode must also pass checks on bottle lifting, position and orientation, return to the initial joint configuration, and gripper aperture. Fast-WAM succeeds in 1 episode (2\%), and ReRead in 25 episodes (50\%).

\paragraph{Real robot.}
The experiments use a Flexiv dual-arm robot. We train Fast-WAM and ReRead for 20K steps each using 200 demonstrations obtained through teleoperation and expert trajectory generation. The clean and random settings correspond to those in simulation.

Each model is evaluated for 25 episodes in each setting, 50 episodes in total, with a limit of 1600 steps per episode. An episode succeeds if the robot picks up the bottle, returns it upright to its original position, and returns the arm to its initial pose within this limit. Fast-WAM fails in all 50 episodes; ReRead succeeds in 18 episodes (36\%).

\subsection{Targeted Intervention on Phase History}
\label{sec:appendix_phase_intervention}

We use the frozen Fast-WAM and ReRead weights after each has been trained for 2K steps on the bottle task, with $D$ denoising steps, replanning after $S$ executed steps, and action chunk length $H$. The evaluation decisions come from 8 expert trajectories whose seeds do not overlap with those of the 200 training demonstrations; replacement trajectories come from the training set. At decision anchors spaced $S$ steps apart, we match pre-grasp approach decisions with post-place retreat decisions, requiring similar images, gripper states, and bottle states. This yields 7 scenes and 14 decisions.

Cross-trajectory replacements use the same active arm, bottle type, clean/random setting, and text instruction as the evaluation decision. Replacement trajectories are ranked by similarity in current pose and bottle position, with preference for trajectories whose past candidate MSE ratio between the pre-grasp and post-place phases is at most 2. We also swap the caches of the two phases within all 7 scenes and reuse these caches at both evaluation decisions of the same trajectory.

The directional metric uses the arm joints. For decision $w$, the expert action of the current phase is assigned coordinate 0 and that of the other phase coordinate 1. The action coordinate along the phase direction and the directional shift after history replacement are defined as
\begin{equation}
\begin{aligned}
    c_w(A)&=\frac{\langle A-A_w^\star,d_w\rangle}{\|d_w\|_2^2},\\
    \Delta c_w&=c_w(\widehat A_w^{\rm opp})-c_w(\widehat A_w^{\rm same}),
\end{aligned}
\end{equation}
where $d_w$ is the reference direction obtained by translating the other phase's expert action displacement to the starting point of the evaluation decision.

\begin{table}[!htbp]
\centering
\caption{Action MSE on phase-ambiguous scenes ($10^{-4}$).}
\label{tab:bottle_phase_mse}
\small
\setlength{\tabcolsep}{7pt}
\begin{tabular}{@{}lr@{}}
\toprule
History condition & Action MSE\\
\midrule
Native candidate & 379.79\\
True history & 149.73\\
Current only & 250.64\\
Cross-trajectory, same phase & 127.52\\
Cross-trajectory, opposite phase & 521.52\\
Same-scene, opposite phase & 519.45\\
\bottomrule
\end{tabular}
\end{table}

As shown in Tab.~\ref{tab:bottle_phase_mse}, true history reduces the action MSE by 40.26\% relative to the current-only readout; opposite-phase history from other trajectories increases the error 4.09-fold relative to same-phase history.

\end{document}